\documentclass[letterpaper, 10 pt, journal, twoside]{ieeetran} 
\IEEEoverridecommandlockouts                              

\usepackage{graphics} 
\usepackage{amsmath} 
\usepackage{amssymb}  
\usepackage{siunitx}
\usepackage{mathtools}
\usepackage{booktabs}
\usepackage[ruled]{algorithm2e}
\usepackage{subcaption} 
\usepackage[bottom=45pt,top=57pt, left=48pt, right=48pt]{geometry}   

\usepackage{adjustbox}
\usepackage{pbox}
\usepackage{dblfloatfix}
\usepackage{graphicx}
\usepackage{xcolor}
\usepackage{hyperref}
\usepackage{caption}
\usepackage{subcaption}
\usepackage{cite}

\begin{document}

\title{
A Unified Approach for Autonomous Volumetric Exploration of Large Scale Environments under Severe Odometry Drift
}


\author{Lukas Schmid$^{1\ast}$, Victor Reijgwart$^{1\ast}$, Lionel Ott$^1$, Juan Nieto$^1$, Roland Siegwart$^1$, and Cesar Cadena$^1$

\thanks{$^\ast$ Authors contributed equally.}%
\thanks{This work was supported by funding from the Microsoft Swiss Joint Research Center, the National Center of Competence in Research (NCCR) Robotics through the SwissNational Science Foundation, and the European Union’s Horizon 2020 research and innovation programme under grant agreement No 101017008.}
\thanks{$^1$ The authors are with Autonomous Systems Lab, Department of Mechanical and Process Engineering, ETH Z\"urich, 8092 Z\"urich, Switzerland
{\tt\small \{schmluk, victorr\}@mavt.ethz.ch}}%
\thanks{Digital Object Identifier (DOI): see top of this page.}%
}

\maketitle

\begin{abstract}
Exploration is a fundamental problem in robot autonomy. 
A major limitation, however, is that during exploration robots oftentimes have to rely on on-board systems alone for state estimation, accumulating significant drift over time in large environments. 
Drift can be detrimental to robot safety and exploration performance.
In this work, a submap-based, multi-layer approach for both mapping and planning is proposed to enable safe and efficient volumetric exploration of large scale environments despite odometry drift.
The central idea of our approach combines local (temporally and spatially) and global mapping to guarantee safety and efficiency. 
Similarly, our planning approach leverages the presented map to compute global volumetric frontiers in a changing global map and utilizes the nature of exploration dealing with partial information for efficient local and global planning.
The presented system is thoroughly evaluated and shown to outperform state of the art methods even under drift-free conditions.
Our system, termed \emph{GLocal}, is made available open source.
\end{abstract}


\section{Introduction}\label{sec:introduction}
\IEEEPARstart{T}{he} ability to autonomously explore unknown environments is a crucial component to robot autonomy and prerequisite for a wide range of applications, ranging from consumer and service robotics to surveying and search and rescue.
In many of these applications, the goal of exploration is to actively map an unknown environment, such that it is fully covered in as little time as possible, and to do so safely.
This is a challenging problem for a number of reasons.
First, due to the nature of exploration, robots have to operate based on the limited information available at each time step. 
A second challenge lies in the conflicting objectives of speed and coverage.
The latter is thus commonly addressed by splitting exploration into the sub-problems of local and global exploration, aiming to maximize exploration speed and coverage, respectively~\cite{history_nbvp,Selin_nbv_fron,dang2019gbplanner, meng_2stage_expl,charrow_ITplanning}.

In local exploration, robots aim to quickly uncover as much of the yet unmapped space as possible while avoiding obstacles.
In contrast to this, global planners track the boundaries between observed and unknown space in the global map to identify promising targets when the local planner gets stuck in a local minimum such as a dead end. 
To address the problem of safety, the majority of exploration approaches use volumetric maps, based on occupancy \cite{hornung2013octomap} or Truncated Signed Distance Fields (TSDF) \cite{oleynikova2017voxblox}, to represent the environment with sufficient detail. 
However, such approaches also make the limiting assumption that perfect pose estimation is available at all times.

In practice, however, exploration is oftentimes desirable in GNSS-denied environments, where robots have to rely on on-board sensing to incrementally estimate their pose, accumulating significant drift during extended missions in large environments.
This results in corruptions in the map that can prevent robots from exploring scenes completely, efficiently, and without collisions. 
Thus, in the context of exploration subject to odometry drift, several additional restrictions are imposed.

\begin{figure}[]
\centering
\includegraphics[width=\linewidth]{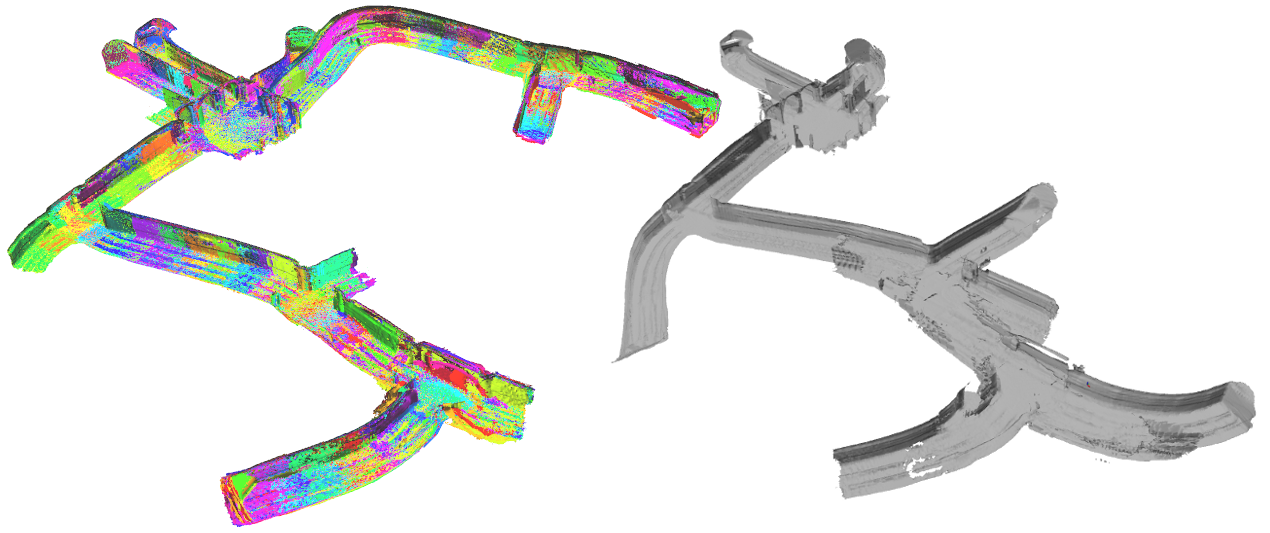} 
\caption{Qualitative comparison of the maps obtained after 20 minutes of autonomous exploration in the \emph{Tunnels} environment at \emph{Moderate} odometry drift. Left: Our proposed system, \emph{GLocal}, utilizes a submap-based (colors) multi-layer planning and mapping approach to explore the tunnels safely despite drift. Right: The approach of Active3D is not able to counteract the accumulation of drift. This leads to corruptions in the map and limits its performance, ultimately causing it to collide with the environment.}
\label{fig:tunnel_qualitative}
\vspace{-6mm}
\end{figure}

Simultaneous Localization And Mapping (SLAM) systems profit from loop closure measurements to refine past pose estimates. Therefore, the global map representation needs the flexibility to accommodate such pose updates.
Although several extensions to rigid grids have been proposed~\cite{reijgwart2020voxgraph, Ho2018Submaps}, the cost of querying these maps scales linearly with time at best, rendering large-scale exploration intractable.
In addition, while a consistent representation of the global map is required to ensure full coverage, local consistency, which often can not be guaranteed by these approaches, is paramount to safety when the robot pose has drifted.

Similarly, planning systems are required to generate safe and informative paths quickly. 
This includes the computation of paths for rapid exploration as well as computation of global goals in a changing map, which can be prohibitively expensive in large volumetric maps.

To overcome these limitations, we propose a multi-layer approach that can provide the different scales required for mapping and planning.
To leverage the advantages of dense volumetric maps in the challenging case of exploration subject to odometry drift, a submap-based approach is presented, producing a temporally local map, a spatially local map, and a global submap collection to incorporate past pose corrections.
Simultaneously, we show how this map representation can be leveraged for efficient and safe exploration planning. 
This includes a local planner that can operate reliably and efficiently in spite of map corruptions due to drift, as well as a submap-based approach for fast, global, volumetric frontier computation.
We show in thorough experimental evaluation that the presented approach is safe and robust when exposed to drift levels common to state of the art odometry systems while simultaneously improving exploration speed also in the absence of drift.
We make the following contributions:
\begin{itemize}
    \item A multi-layer architecture for both planning and mapping, enabling efficient local and global volumetric exploration and safe navigation in large environments under severe odometry drift.
    \item A submap-based approach to efficiently compute and track global frontiers in volumetric submap-based global maps that can deform to accommodate past pose corrections.
    \item Thorough evaluations of the presented system in extensive experiments. We make both our framework, termed \emph{GLocal}\footnote{\url{https://github.com/ethz-asl/glocal_exploration}.}, and the simulation setup\footnote{\label{footnote:unreal_airsim}\url{https://github.com/ethz-asl/unreal_airsim}} available as open source for the benefit of the community.
\end{itemize}

\section{Related Work}

\subsection{Exploration Planning}
Exploration is a fundamental problem in robotics and has attracted considerable research interest. 
Although there exists a variety of approaches~\cite{aut_expl_comparison}, the majority can be grouped into frontier and sampling-based methods.

Frontier-based exploration~\cite{yamauchi1997frontier} explicitly computes the boundary between observed and unobserved space to select goals for exploration. 
This has the advantage of only terminating once all boundaries of the map are waterproof surfaces, thus guaranteeing full coverage. 
Although extensions for rapid flight exist~\cite{cieslewski_2017}, frontier computation can be expensive in large volumetric maps~\cite{Keidar2014frontier} and is not directly transferable to different tasks.

Sampling-based approaches sample viewpoints or paths, e. g. Rapidly-exploring Random Trees~\cite{Selin_nbv_fron, nbvp_object_search, nbvp_bircher}, and select the most promising candidates based on a utility estimate. 
This permits a broad range of utility objectives~\cite{nbvp_object_search, Schmid20ActivePlanning, song_3d_rec}, therefore allowing more fine-grained optimization of the local robot trajectory. 
However, coverage depends on the sampling scheme and planning horizon, which can make these methods prone to local minima.

Recent methods thus oftentimes support a sampling-based local planner with additional global information to improve coverage.
\cite{history_nbvp, Selin_nbv_fron} utilize a Receding Horizon (RH)-RRT~\cite{nbvp_bircher} for local exploration and store global information by sampling an additional history graph~\cite{history_nbvp} or storing the gain of non-executed viewpoints in a Gaussian Process (GP)\cite{Selin_nbv_fron}.
Dang \textit{et al.} \cite{dang2019gbplanner} extend \cite{nbvp_bircher} to a local graph of shortest paths that is added to a global graph, which is queried when local gains are low.

Corah \textit{et al.}~\cite{gmm_multirobot_expl} also store global information in a library of views, but use motion primitives for local exploration. 
However, instead of decoupling local from global exploration, the distance to a view from the global library is added to the gain to encourage paths toward informative areas of the map. 
A fully coupled approach is presented in \cite{Schmid20ActivePlanning}, where a single RRT* tree is expanded and maintained for unified local exploration and global outreach.

A third family of methods explicitly computes frontiers and uses sampling to enhance the found paths.
Charrow et al. \cite{charrow_ITplanning} supplement paths to global frontiers with local motion primitives.
Alternatively, \cite{meng_2stage_expl, dai2020fast} use frontiers to sample candidate viewpoints, and cast them into a traveling salesman problem to search an optimal visitation sequence \cite{meng_2stage_expl} or compute a utility for each candidate \cite{dai2020fast}.

However, these approaches do not yet fully embrace the difference in objectives between local and global planning.
We argue that the majority of the relevant information is generally uncovered in the next moves of the robot and propose a local planner focused on a minimal temporal and spatial horizon to guarantee safety and consistency in spite of drift, without sacrificing exploration speed.
To guarantee coverage, it is combined with explicit global frontier computation in a map that deforms as its past pose estimates improve.

\subsection{Mapping for Exploration}
State-of-the-art exploration approaches typically employ volumetric maps due to their effectiveness at representing observedness and traversability at all points in space. A traditional volumetric map representation is occupancy \cite{hornung2013octomap}. Recently, Signed Distance Fields \cite{izadi_kinectfusion_2011,oleynikova2017voxblox} have gained traction given their additional benefits for tasks including motion planning \cite{ratliff_chomp_2009,oleynikova2016continuous-time}. Most volumetric mapping frameworks, however, integrate all measurements into a single rigid grid which cannot readily be updated when corrected past pose estimates become available.
Several extensions have been proposed that overcome this limitation, e.g. by modelling the map as a collection of submaps which can be aligned through pose graph optimization \cite{hess_real-time_2016,millane_c-blox_2018,Fioraio2015Subvolumes,reijgwart2020voxgraph}.

\subsection{Frontier Detection}
The goal of frontier detection is to identify the boundaries between known and unknown parts of the map and cluster them such that they can act as navigation goals. However, this operation can be prohibitively expensive in large volumetric maps.
In their seminal work, Keidar and Kaminka \cite{Keidar2014frontier} propose two methods for efficient frontier computation. In Wave Front Detection (WFD), two breadth-first searches are run to identify and cluster frontiers, respectively. In Fast Frontier Detection (FFD), the contour of each scan is computed and then used to update the frontiers of the map.
Nonetheless, this requires processing of the full map in case of WFD or redundant integration after every sensor measurement in the case of FFD, which is not readily applicable to deformable volumetric global maps.
To overcome both problems, we leverage the submap structure of our global map and propose a submap-based approach for efficient global frontier computation in deforming maps.
In recent work, Orsulic \textit{et al.}~\cite{Orsulic2019frontier} independently developed the concept of submap frontiers for 2D occupancy submaps. 
However, only the detection of frontiers is discussed without extensions for planning or reachability guarantees. 
We thus see this work as an extension of the concept to volumetric 3D maps and propose an efficient spatial hashing approach for clustering and goal selection in planning.

\subsection{Exploration subject to Odometry Drift}
To address exploration subject to odometry drift, 
\cite{Akdeniz2015Topological} propose a topological graph of bubble surfaces, which is recursively extended and explored until place recognition occurs. 
In similar fashion to our approach, Cieslewski \textit{et al.}~\cite{cieslewski2019exploration} also take inspiration from graph SLAM and build a graph of local 2D polygons for every measurement. Without a need for global consistency, the local consistency idea is used to consolidate frontiers based on proximity in the topological graph.
A volumetric approach is presented by Ho \textit{et al.}~\cite{Ho2018Submaps}, providing a general submap interface that could potentially be used for exploration. 
However, all lookups scale linearly with the number of submaps, which quickly becomes intractable in large environments.

\section{System Overview}

\begin{figure}
\centering
    \begin{subfigure}{\linewidth}
        \includegraphics[width=0.95\linewidth]{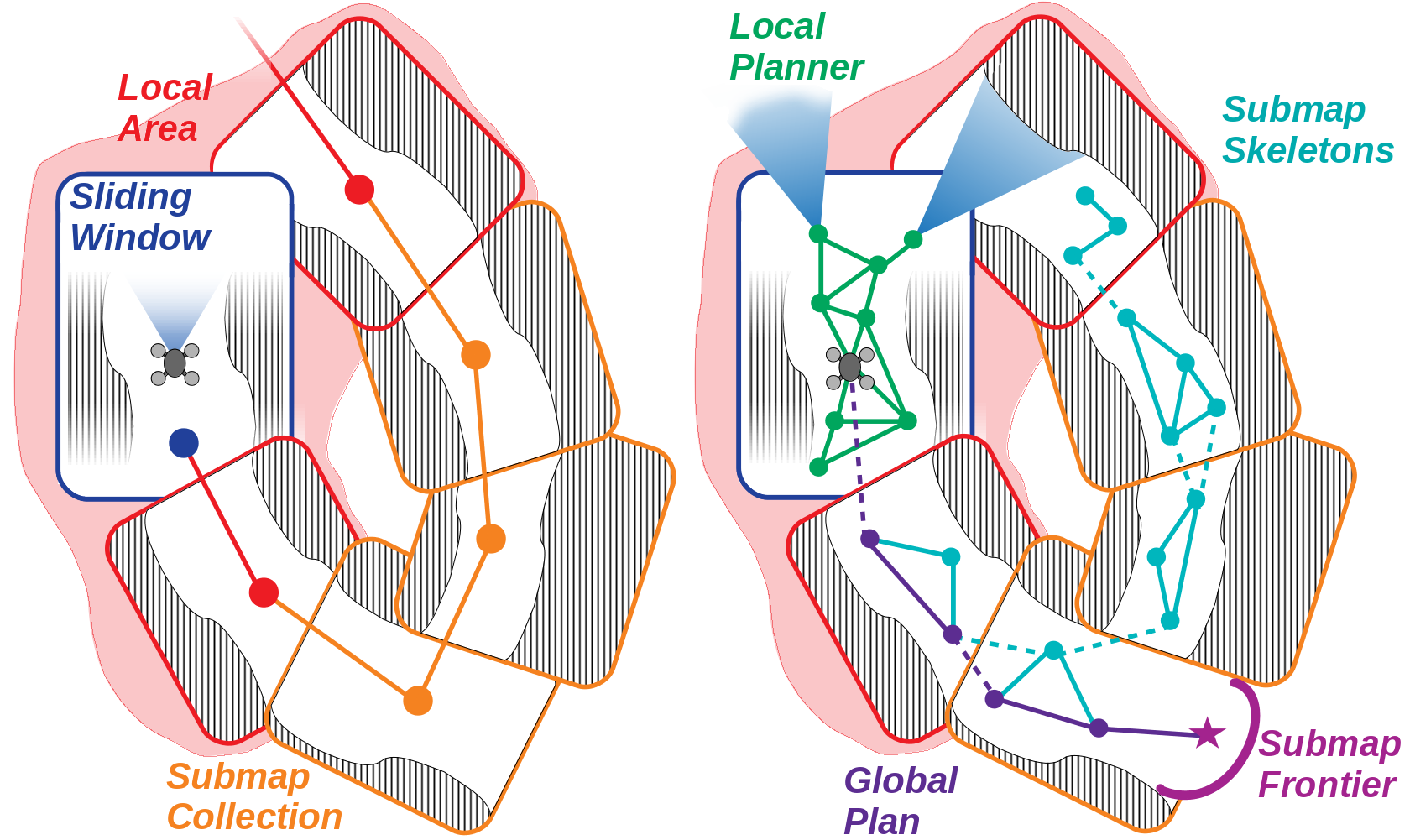}
        \caption{Graphical outline of our mapping (left) and planning (right) approach.}
        \label{fig:system_overview_mapping}
    \end{subfigure} 
    \\
    \begin{subfigure}{\linewidth}
        \centering
        \includegraphics[width=0.95\linewidth]{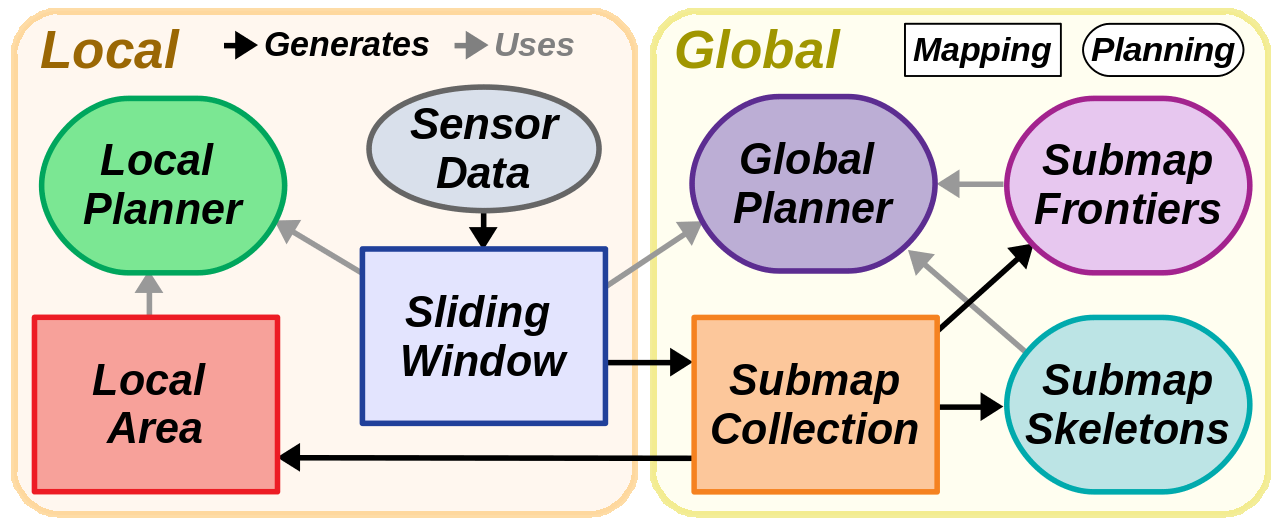}
        \caption{Schematic overview of the presented system.}
        \label{fig:system_overview_schematic}
    \end{subfigure} 
        \vspace{-1.5mm}
    \caption{System overview. Sensor data is integrated into a temporally local sliding window map (blue). Periodically, the sliding window is stored as submaps together with a pose graph to represent the global map (orange). Spatially close submaps are combined in the local area (red). The local planner samples a graph in the sliding window for safety (green) and utilizes the local area to efficiently compute gains (light blue). Upon completion, each submap is skeletonized (teal) and stores frontier candidates (pink) to efficiently compute global plans (purple) at any configuration of submaps.}
    \label{fig:system_overview}
    \vspace{-6mm}
\end{figure}

The task of exploration subject to odometry drift imposes distinct requirements on both the map and the planner. 
The map needs the flexibility to accommodate past pose corrections and robustness w.r.t. corruptions due to odometry drift. 
The planner requires safe and efficient local map lookups and global coverage. 
Thus, the central idea of our approach is to use multiple layers of both mapping and planning to provide the different scales required. 
This includes a global, a temporally local, and a spatially local map, as well as a local planner for safe, reliable and quick exploration and a frontier-based global planner to guarantee full coverage.
An overview of the proposed system is given in figure \ref{fig:system_overview} and each component is described in more detail below.

\subsection{Sliding Window}
An important role of the map in path planning is to avoid collision. 
However, with imperfect state estimation, this can not be guaranteed when a monolithic global map is used. 
We thus propose a temporally local map as the first layer of mapping, termed the \emph{sliding window} map, where, under the reasonable assumption that drift is low over a short time horizon, sensor data can be accumulated into a local TSDF grid \cite{oleynikova2017voxblox}.
In a sliding window fashion, measurements expire once the relative uncertainty between the pose at which they were recorded and the current pose exceeds a threshold. 
They are then removed from the map by reversing their TSDF integration step. 

This guarantees that the sliding window contains only up-to-date information that can be used by all planners for reliable collision checking.
Because sensor data typically arrives at a uniform rate, the sliding window has constant bounds for memory and CPU consumption.

\subsection{Global Map}
Extending the local consistency assumption, the sliding window map is periodically stored as submaps, which together represent the \emph{global map}. We employ voxgraph \cite{reijgwart2020voxgraph} to organize all submaps in a pose graph that connects subsequent submaps with odometry priors and surface alignment edges, allowing to optimize for a globally consistent alignment of submaps. 

\subsection{Local Area}
For a local planner to compute meaningful information gains, it requires data about its environment, such as whether a place was observed at an earlier time. 
In large scale environments, only few submaps contribute to local planning, thus the last layer in our mapping approach is a spatially local map, the \emph{local area}. 
It consists of a single TSDF map that contains all submaps that overlap with the sliding window.
Existing submaps are integrated and de-integrated when they enter or leave the proximity of the sliding window, respectively. 
Although maintaining the local area may be more expensive than directly querying the global map early on, it enables $O(1)$ lookup independent of the map size or mission duration for consistent and reliable operation of the local planner.

\subsection{Local Planner}
To maximize flexibility and performance of the local path, we sample viewpoints $v$ in the sliding window and connect a set of edges $E(v) = \{e(v,v')\}_{v' \in N(v)}$ to all nearby viewpoints $N(v)$, forming a graph $\mathcal{G} = \{\mathcal{V}, \mathcal{E}\}$. 
We adopt the methodology of \cite{Schmid20ActivePlanning} and compute a gain $g(v)$ for each viewpoint $v$ in the local area and a cost $c(e)$ for each edge $e$.
We use the number of unobserved voxels as $g(v)$ and the edge length as $c(e)$. 

Once execution of an edge is finished, all colliding edges are removed to guarantee safety and disconnected viewpoints are pruned from the graph. 
Each viewpoint is then assigned an \emph{active edge} $a(v) \in E(v)$, such that the set of active edges forms a tree starting from the current robot pose. 
These are updated by computing the global normalized value \cite{Schmid20ActivePlanning}, i.e. the maximum of the ratio between accumulated gain and accumulated cost anywhere in the tree.
\begin{gather*}
\vspace{-3mm}
    a(v)^{(i+1)} = \arg \max_{e \in E(v)}
        \max_{v_j \in \text{subtree}(e)}
        \frac{
            \sum_{v_k \in path(v_j)}g(v_k)
        }{
            \sum_{v_k \in \text{path}(v_j)}c(a^{(i)}(v_k))
        } 
    \\
    \text{path}(v) = \{v, a^{(i)}(v), a^{(i)}(a^{(i)}(v)), \dots , \text{root}\}
    \label{eq:value_comp}
\end{gather*}

This process is iterated until convergence to find the most informative path in the graph and execute the first edge thereof. 
Afterwards all viewpoint gains are updated since they, by construction, lie in the currently changing map.

The bounded extent of the sliding window and constant complexity of gain computation, which dominates local planning cost, result in a graph of approximately constant density moving along with the sliding window for reliable and efficient local exploration.
Once none of the nodes in the local graph return significant gain, the system switches to global planning to identify a new exploration site.

\subsection{Submap Frontiers}
Because the global map consists of submaps which can move with respect to each other, global frontiers can disappear or re-appear in a given submap. Since global volumetric frontier search is an expensive operation \cite{Keidar2014frontier} and would require rebuilding the global map for every configuration of submaps, we propose \emph{submap frontiers} as an efficient way to compute and track all global frontier points. 

Similar to \cite{cieslewski2019exploration}, the central idea is to leverage the local consistency of each submap to reduce the search space of possible frontier points, such that they can be efficiently detected at any configuration of the global map.

We define a frontier as an unknown cell that neighbors at least one observed free cell, and attribute ownership of that frontier to the submap that contains the observed cell.
Because observed cells can never become unobserved through superposition with another submap, it is sufficient to focus solely on the unobserved cells of a submap. 
Similarly, because submaps are frozen once created, any cells that become observed through superposition must be from other submaps. 
Therefore, all points that could potentially be a frontier owned by a submap are unknown cells next to free space cells of that submap, i.e. its frontiers.
By computing and caching frontier candidates upon submap completion, the search space is reduced from volumetric to a subset of the surface of a submap, reducing the number of candidates by two orders of magnitude in our experiments.

To identify global frontiers, candidates are transformed to global coordinates and aggregated in a 3D spatial hash set, removing duplicates.
Inactive frontiers are removed by checking whether the unobserved cell is observed in any submap.
Eventually, a clustering into connected active frontiers can be computed using region growing along the hash set in $O(N_{active\_candidates})$.

\subsection{Global Planner}
Because even a small frontier can lead to potentially large unknown spaces, we select the closest frontier as a goal.
Using the Euclidean distance as a lower bound for the path length, we compute paths to the closest goal and iteratively prune targets located further than the current shortest path.

To compute feasible paths efficiently, we again make use of the submap structure. Upon completion, submaps are skeletonized \cite{oleynikova2018sparse}, i.e. a traversability graph is computed and stored with the submap.
At query time the skeleton graphs of overlapping submaps are connected and a feasible path can be found using, for instance, the A* algorithm \cite{hart1968formal}.

To guarantee safety also during global path execution and improve performance, we employ path shortening based on the sliding window map.


\section{Evaluations}

\begin{figure}
\centering
    \begin{subfigure}{0.53\linewidth}
        \includegraphics[width=\linewidth]{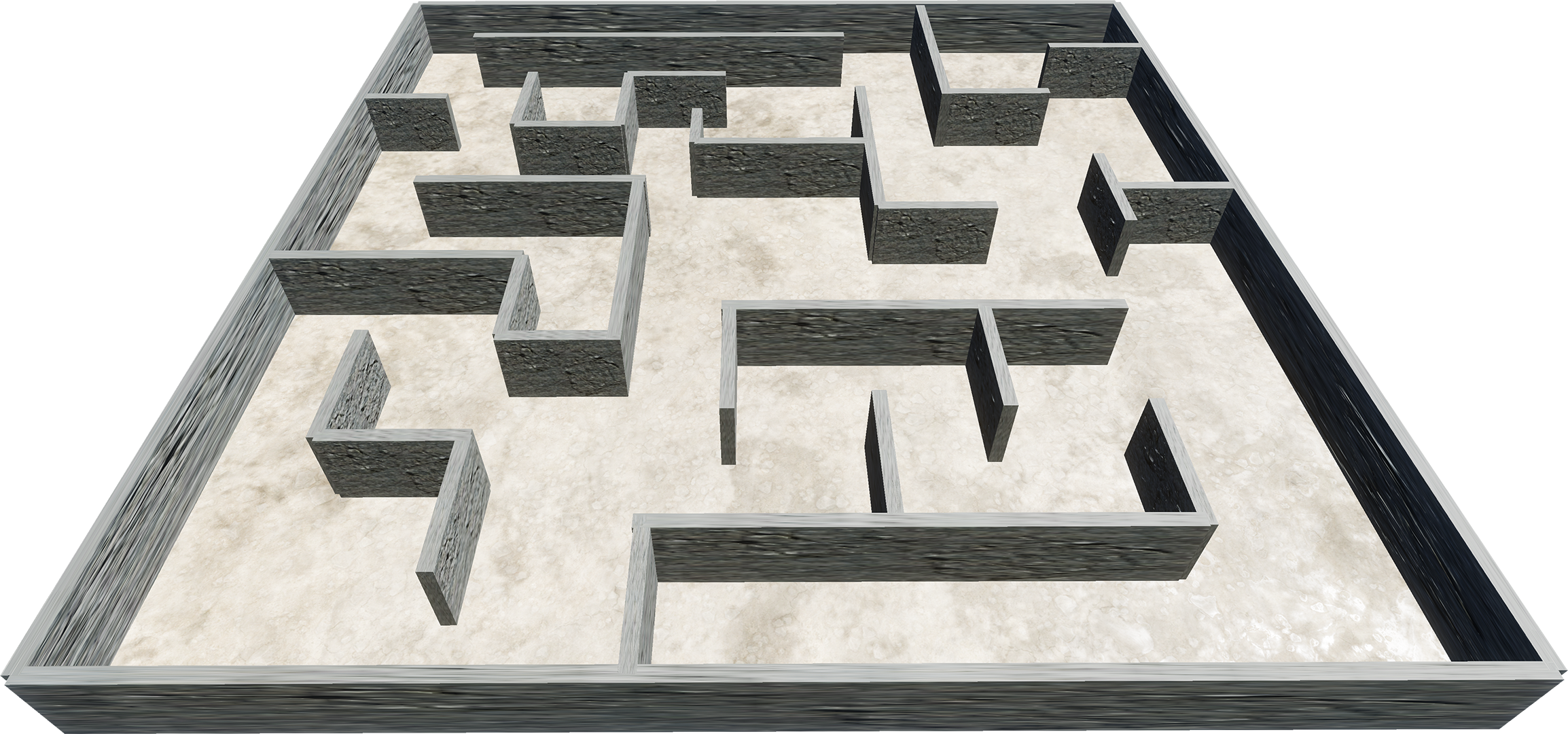}
        \label{fig:unreal_maze_overview}
    \end{subfigure} 
    \begin{subfigure}{0.45\linewidth}
        \includegraphics[width=\linewidth]{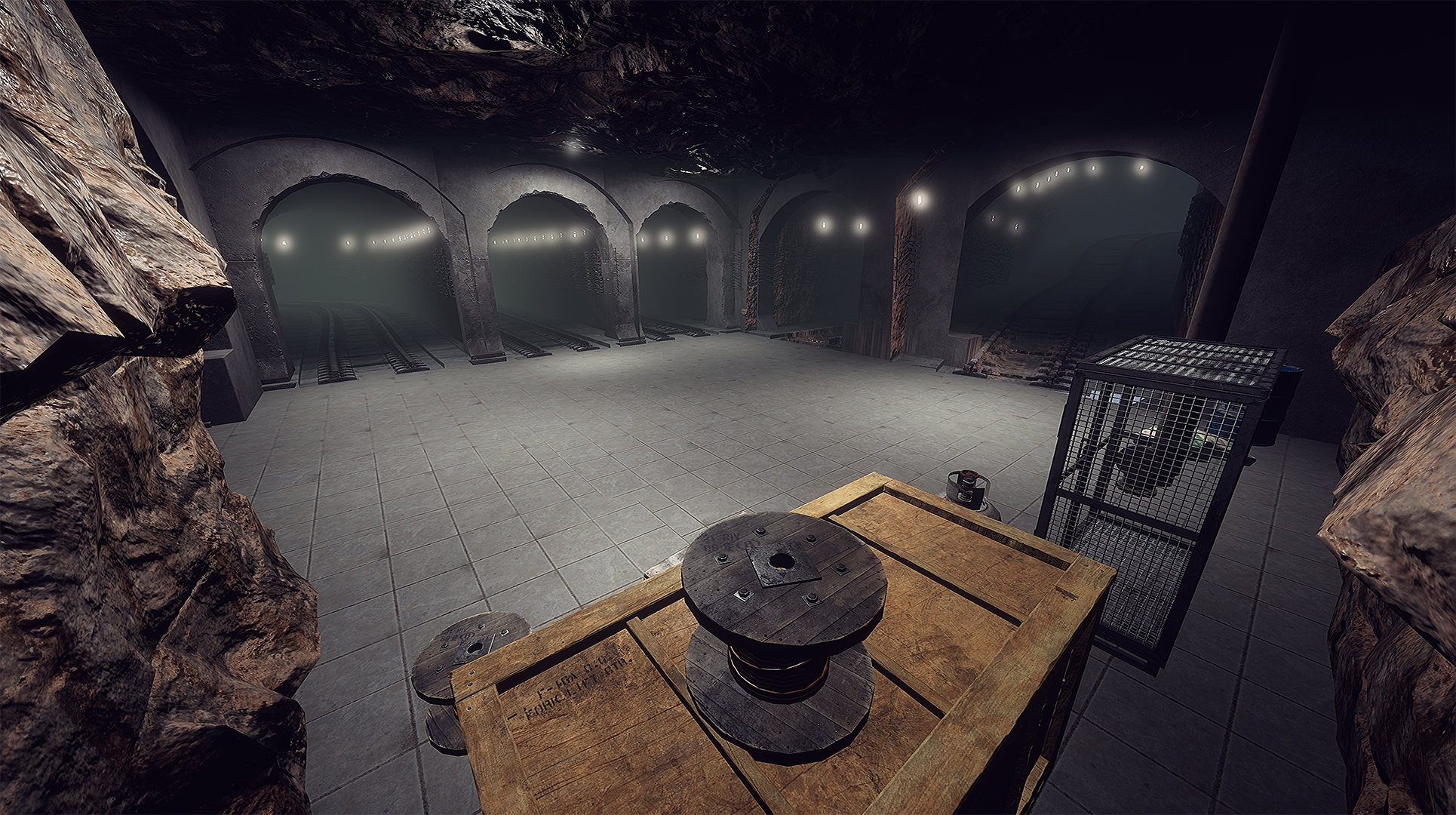}
        \label{fig:unreal_tunnel_detail}
    \end{subfigure} 
    \vspace{-5mm}
    \caption{Simulation experiments are carried out in two photo-realistic environments using Unreal Engine. An overview of the \emph{Maze} is shown on the left, followed by a detail shot of the \emph{Tunnels} on the right. For an overview of the layout of the  \emph{Tunnels}, see Fig.~\ref{fig:tunnel_qualitative}.}
    \label{fig:unreal_environments}
    \vspace{-5mm}
\end{figure}

We evaluate our proposed system in extensive life-like simulations using Unreal Engine\footnote{\url{https://www.unrealengine.com/en-US/}}, a high-fidelity game engine capable of modelling complex and photo-realistic environments.
To model the MAV dynamics, sensing, and flight controller, we use Microsoft Airsim \cite{shah2018airsim}. A custom ROS interface is used to connect it to the rest of our software stack and simulate odometry drift\textsuperscript{\ref{footnote:unreal_airsim}}. 
Experiments are carried out in the challenging and narrow \emph{Maze} environment of \cite{Schmid20ActivePlanning} of $40\times40\times3 \si{m}$ and a large scale underground \emph{Tunnels} environment spanning $123\times106\times10 \si{m}$, shown in Fig. \ref{fig:unreal_environments}.

We compare our method, termed \emph{GLocal}, against the mav\_active\_3d\_planner introduced in \cite{Schmid20ActivePlanning}, further referred to as \emph{Active3D}.
It expands and rewires a single large tree and focuses on performance by optimizing the utility of the path in global context.
We further compare against \emph{GBPlanner} \cite{dang2019gbplanner}, a graph-based, local-global exploration algorithm that prioritizes safety of the planned paths by favoring slightly sub-optimal paths that are far from obstacles.

For each setting, 10 experiments were conducted and means and standard deviations are reported. Identical configurations of the MAV as shown in Tab.~\ref{tab:params} were used for all experiments.
The MAV always starts in the same configuration at the center of the simulation world.

\subsection{Exploration Performance}

\begin{figure}
     \centering
     \includegraphics[width=\linewidth]{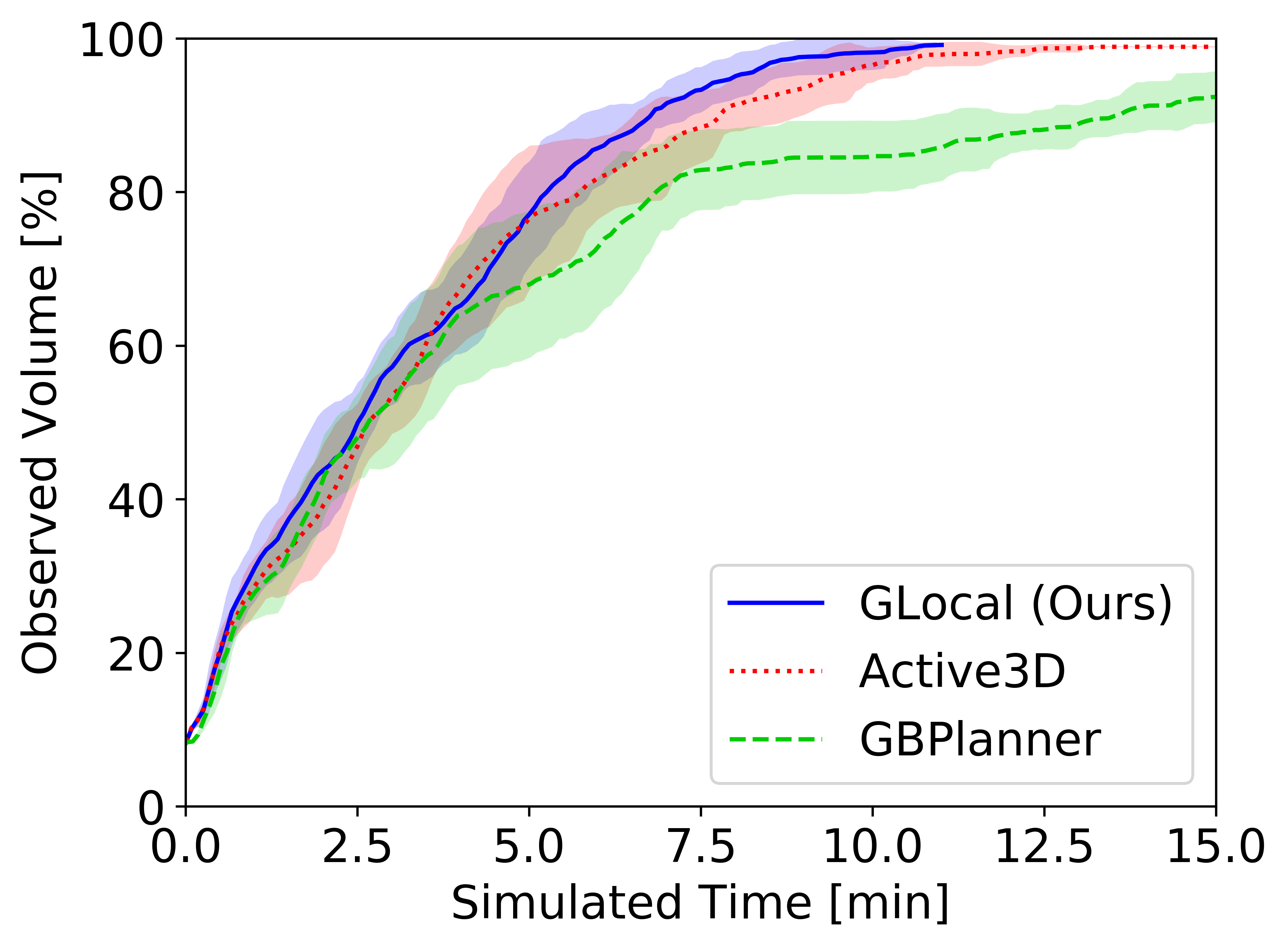}
     \caption{Exploration progress over time in the Maze without any drift, as mean and standard deviation over 10 experiments. Although prioritizing safety, GLocal matches the other exploration algorithms even in the absence of drift. Due to the explicit global frontier representation, GLocal terminates within at most $11$ minutes, signifying a speed-up of $19.5\%$ on average over Active3D.}
     \label{fig:no_drift}
     \vspace{-5mm}
\end{figure}

To evaluate the performance of our system, exploration progress over time in the Maze without any drift is reported in Fig.~\ref{fig:no_drift}. 
Local planning performance is reflected in the first 5 minutes of the graph. 
Most notably, our method shows high exploration speed that is on par with the purely performance oriented Active3D system. 
While Active3D optimizes paths globally with far look-ahead, our system prioritizes reliability and plans paths on a significantly shorter horizon and scale, typically in the range of few seconds and $\sim1\times$ sensing range. 
This observation suggests that our assumption, that local exploration performance is governed primarily by the information within a minimal spatial neighborhood and temporal look-ahead, holds in the Maze scenario.

Second, the importance of the global planner becomes apparent around 5 minutes, when large parts of the maze are already explored.
Here again, the concept of partial information, i.e. the idea that any frontier could lead to arbitrary, potentially large further gains, and the capabilities of GLocal to quickly identify and plan a path to the closest frontier gives it an edge over the gain-based global planners of Active3D and GBPlanner. 
Both Active3D and GBPlanner select global goals based on a trade-off between view-based gain and cost, which can favor further away goals leading to potentially sub-optimal trajectories.
This effect is more pronounced for GBPlanner, whose performance suffers earlier from global planning and is further reduced as more of the map gets explored.
This is likely due to the fact that GBPlanner approximates the path length by the Euclidean distance between the robot and the goal to save compute, potentially degrading the quality of goal selection when this assumption is violated.

An advantage of explicitly computing all frontiers in the map is that exploration can be terminated when no more frontiers are detected, guaranteeing full coverage.
All runs of GLocal consistently explored $98\%$ of the volume enclosing the maze and terminated between $7.58$ and $11.0$ minutes, signifying a speed-up of $19.5\%$ on average over Active3D.  
This shows that GLocal, although optimized for safety and robustness in a changing map, can match or outperform monolithic approaches even when there is no drift.

\begin{table}[t]
    \centering
    \caption{Simulation (top) and system (bottom) parameters used.}
    \begin{tabular}{lrrlrr}
        \toprule
        Max. Velocity & $1$ & $\si{m/s}$ & Max. Duration & $15$ & $\si{min}$ \\
        Lidar FoV & $45$ & $\deg$ & Lidar Range & $10$ & $\si{m}$ \\
        Lidar Frequency & $10$ & $\si{Hz}$ & Lidar Mount Pitch & 15 & $\si{deg}$ \\
        \midrule
        Sliding Window Size & $10$ & $\si{s}$ & Map Resolution & 0.2 & $\si{m}$ \\
        \bottomrule
    \end{tabular}
    
    \label{tab:params}
         \vspace{-2mm}
\end{table}

\subsection{Drift Robustness}

\begin{figure*}[]
\begin{subfigure}{.32\textwidth}
  \centering
  \includegraphics[width=\linewidth]{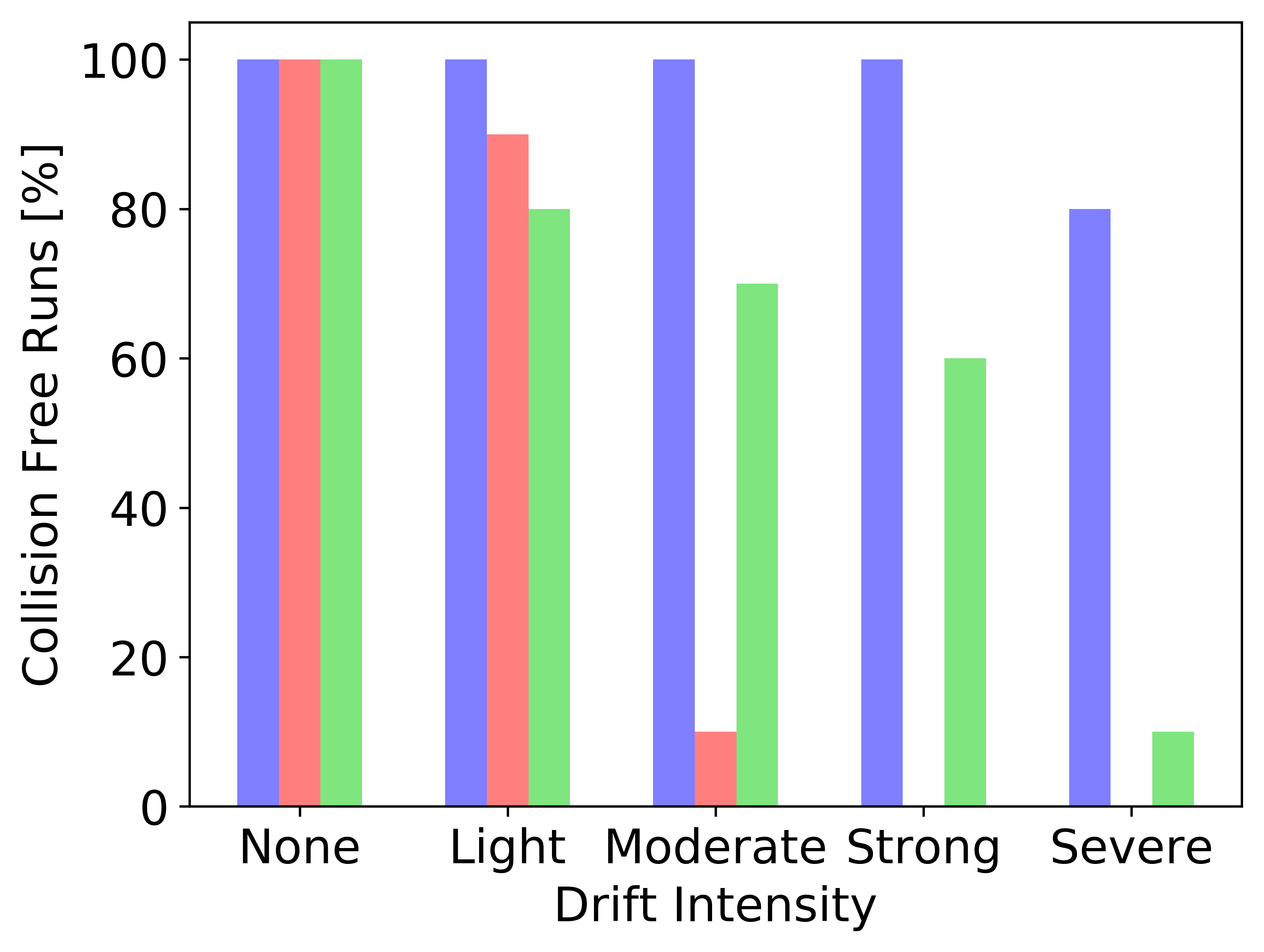}  
  \caption{Percentage of runs without collision.}
  \label{fig:drift1}
\end{subfigure}
\begin{subfigure}{.32\textwidth}
  \centering
  \includegraphics[width=\linewidth]{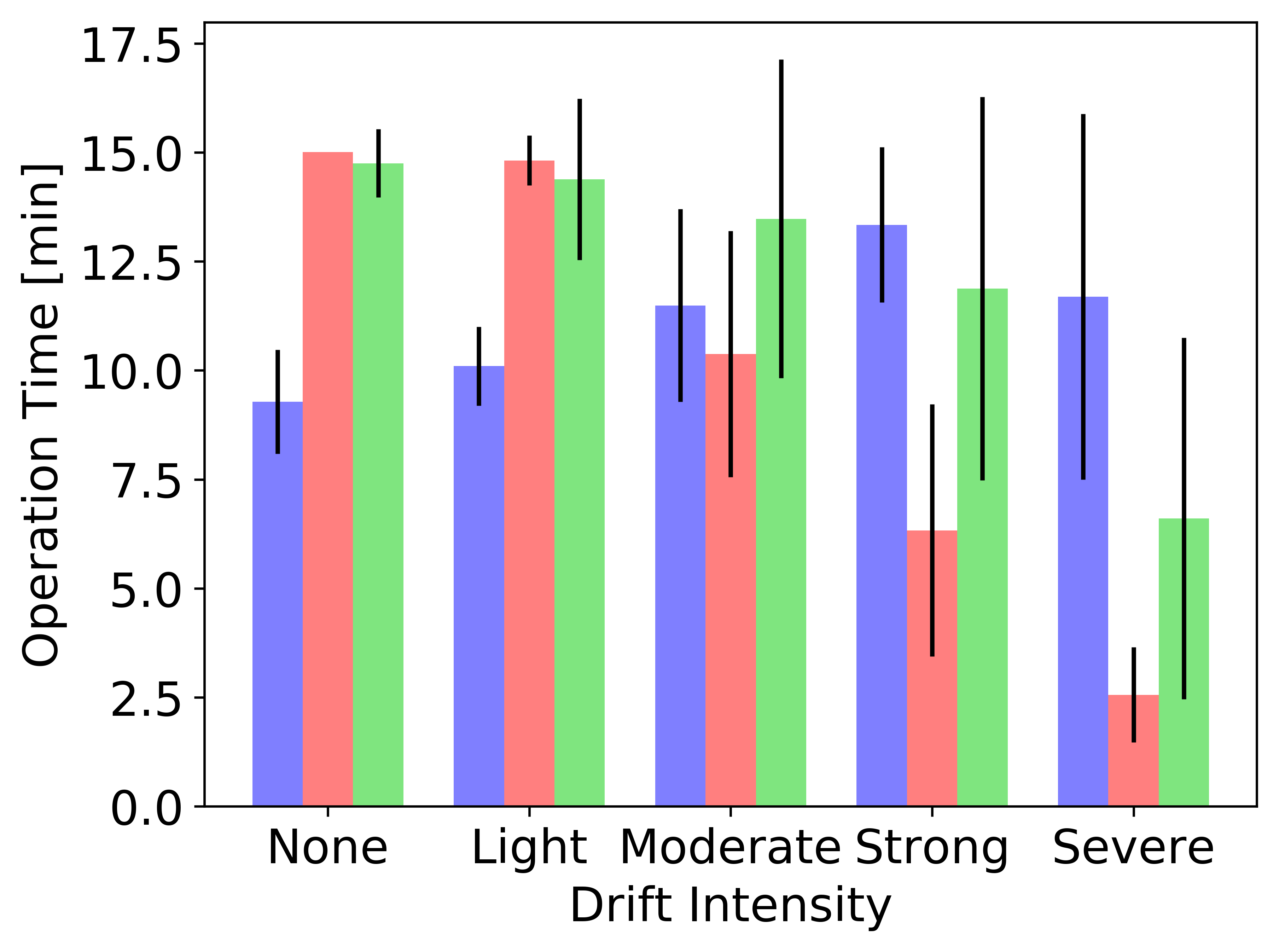}  
  \caption{Total operation time.}
  \label{fig:drift2}
\end{subfigure}
\begin{subfigure}{.32\textwidth}
  \includegraphics[width=\linewidth]{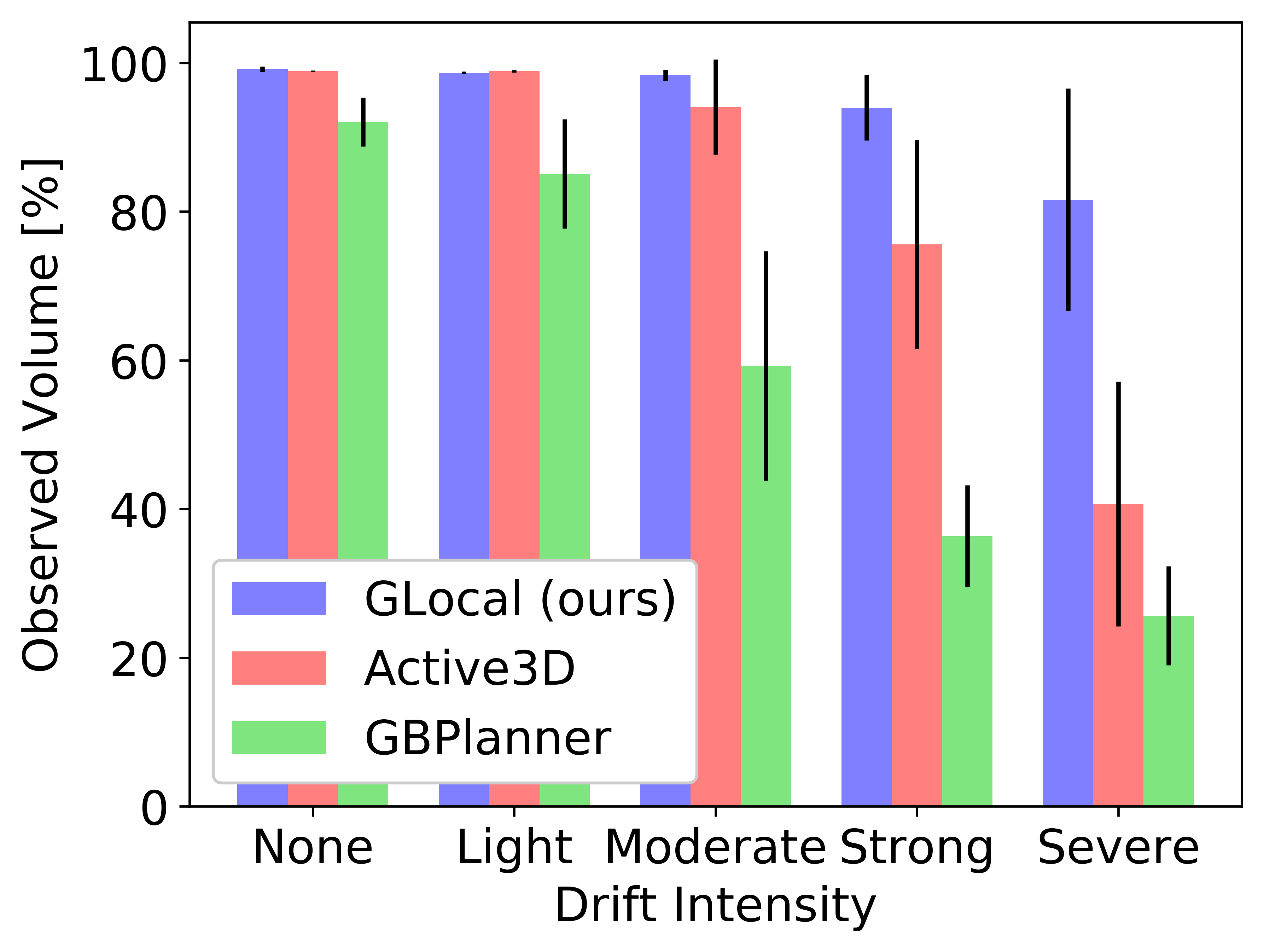}  
  \caption{Final exploration percentage.}
  \label{fig:drift3}
\end{subfigure}
\caption{Exploration in the Maze scenario subject to drift of varying magnitudes (see Tab.~\ref{tab:drift}) over 10 runs. While Active3D and GBPlanner tend to collide earlier and more frequently with increasing levels of drift, GLocal is able to remain safe up until \emph{Severe} drift is reached. In addition, due its multi-layer architecture, it is able to continue exploration despite corruptions in the map.}
\label{fig:drift}
\vspace{-5mm}
\end{figure*}

\begin{table}[t]
    \centering
    \caption{Magnitude of simulated drift per $\SI{100}{m}$ traveled, measured as mean and standard deviation over 10 runs.}
    \resizebox{\columnwidth}{!}{%
    \begin{tabular}{lcccc}
        \toprule
        Drift & Light & Moderate & Strong & Severe \\
        \midrule
        Position [m] & $0.31 \pm0.11$ & $0.56 \pm0.23$ & $1.14 \pm0.74$ & $2.76 \pm1.39$ \\
        Rotation [deg] & $0.95 \pm0.27$ & $2.10 \pm0.42$ & $4.59 \pm1.35$ & $8.85 \pm5.63$\\
        \bottomrule
    \end{tabular}}
    \label{tab:drift}
    \vspace{-5mm}
\end{table}

To verify the robustness of our proposed method with respect to odometry drift, experiments with different magnitudes of drift, as specified in Tab.~\ref{tab:drift},  were conducted in the Maze.
To simulate realistic odometry drift, white Gaussian noise is applied at a low frequency of $0.1\si{Hz}$ to the position and yaw velocity, as pitch and roll are typically well observable using IMUs.
According to a recent study \cite{Delmerico2018drift}, state of the art odometry estimators accumulate $\sim0.6\si{m}$ position and $\sim3\si{deg}$ yaw error per $100\si{m}$, which corresponds to \emph{Moderate} drift in our experiments. 
Drift levels \emph{Strong} and \emph{Severe} were chosen exponentially larger to probe the capabilities of the proposed method.

The results of exploration subject to drift is summarized in Fig.~\ref{fig:drift}. 
Most importantly, Fig.~\ref{fig:drift1} highlights the number of runs that finished without collision. 
Notably, GLocal stays safe in all cases up to \emph{Strong} drift, and even successfully completes $80\%$ of the runs subject to \emph{Severe} drift.
Active3D which utilizes a rigid monolithic map and focuses solely on performance by contrast has high collision rates already for low magnitudes of drift.
Although GBPlanner is also based on a monolithic map, its extra safety considerations allow it to stay collision free more often.
However, such monolithic approaches exhibit a $30-90\%$ collision rate at drift rates typical to state of the art odometry estimators, which poses considerable risks for deployment of such a system.

Similar effects are observed in Fig.~\ref{fig:drift2}, where total operation times are reduced through earlier collision at higher levels of drift, culminating in all runs of Active3D colliding within 4.7 minutes when subject to \emph{Severe} drift.
It is worth pointing out that for GLocal, all runs up to \emph{Strong} drift and part of the latter terminated due to completion and not collision.
Here, the initial increase in completion time highlights the increased demands on the system to compute and track global paths and continue local exploration in spite of map corruptions.

Fig.~\ref{fig:drift3} presents the percentage of the maze explored during operation.
An interesting observation is the discrepancy between operation time and exploration progress at \emph{Strong} and \emph{Severe} levels of drift.
The fact that Active3D explored a larger area than GBPlanner in spite of vastly different operation times indicates that the systems are limited by their ability to deal with map corruptions induced by significant amounts of drift or over longer times.
In contrast to this, GLocal plans local paths solely in the sliding window, allowing it to e.g. traverse a wall in the global map that is misplaced due to drift perfectly safely.
This enables it to continue exploration even under significant displacements which can later be corrected by loop closures.

\subsection{Large Scale Tunnel Exploration}

\begin{figure}[]
\centering
\includegraphics[width=\linewidth]{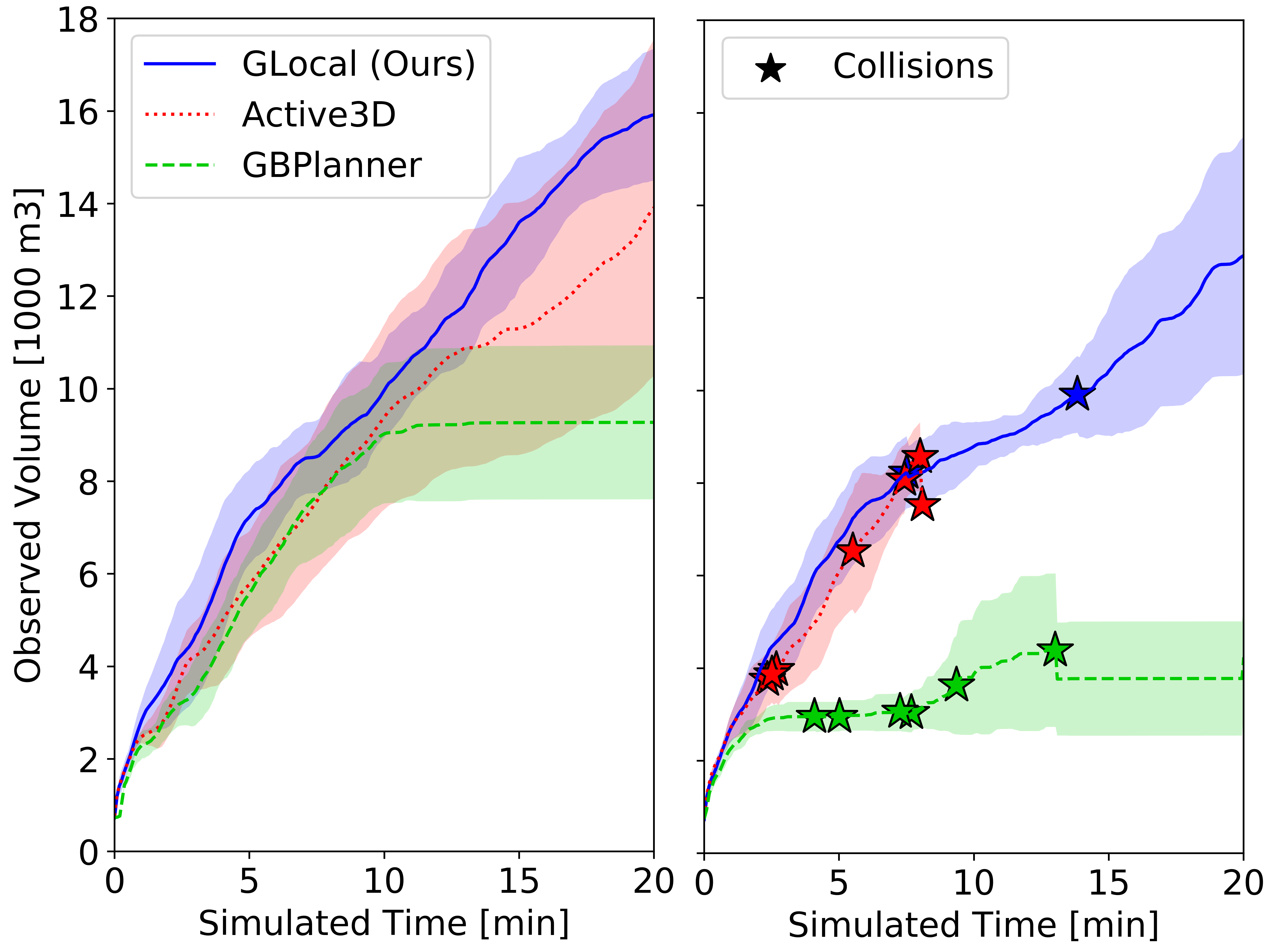} 
\caption{Exploration progress in the \emph{Tunnels} without drift (left) and subject to \emph{Moderate} drift (right) as mean and standard deviation over 10 runs. While all methods perform well in local exploration, the ability to efficiently identify paths to global frontiers is of increased importance in large scale environments. Furthermore, for GLocal only a minor reduction in performance is observed while avoiding collisions when subject to drift.}
\label{fig:tunnel}
\vspace{-5mm}
\end{figure}

To verify the systems capabilities, experiments in an extensive underground environment are performed and reported in Fig.~\ref{fig:tunnel}.
Performance in the absence of drift is highlighted in Fig.~\ref{fig:tunnel}, left.
Similar to the Maze, all methods tend to perform well initially during local exploration.
However, a particular challenge of such subterranean environments lies in the bifurcated layout and length of the tunnels.
This is reflected in the performance once global planning becomes more relevant.
In particular, GBPlanner has difficulty transitioning from one branch of the tunnel system to the other, resulting in only partial exploration.
Similarly, the differences in exploration speed between GLocal and Active3D become more pronounced after 10 minutes.
Here, the ability to identify and reach even small, far away frontiers gives GLocal the edge over its counterpart.

The right half of Fig~\ref{fig:tunnel} presents identical experiments subject to \emph{Moderate} drift.
Since Active3D does not account for safety it does not drop in exploration speed, but tends to crash early in all runs.
Opposite to that, GBPlanner and GLocal prioritize safety and complete $30\%$ and $80\%$ of the runs, respectively.
However, GBPlanner is not able to able to handle map corruptions as well as GLocal, leading to limited exploration progress.

A qualitative comparison of the obtained maps is presented in Fig.~\ref{fig:tunnel_qualitative}.
Notably, the submap-based approach of GLocal gives it the flexibility to align repeated observations when traversing between global goals.
On the other, Active3D is not able to counteract the accumulation of drift.
This leads to corruptions in the map, limiting its performance and safety.

\subsection{Ablation Study}
To investigate the importance of constant-time local planning and efficient frontier detection, an ablation study is performed. 
As counterpart, the general interface of directly querying the submap collection proposed in \cite{Ho2018Submaps} is used for local planning and global frontier search.
The means and standard deviations of 10 runs in the \emph{Maze} at \emph{Moderate} drift are presented in Tab.~\ref{tab:ablation}.
The local planner is of particular importance at the start of exploration, where the methods without the local area tend to fall behind their complements.
Similarly, the latter stages of exploration are typically dominated by global plans, highlighted by the fact that the no-local-area planners are able to overtake the no-submap-frontier planners towards the end of exploration.
Eventually, only the combination of efficient local and global planning is able to achieve best performance.
It is worth pointing out that since \cite{Ho2018Submaps} additionally scales with time, these findings can be expected to be even more pronounced in larger environments.

\begin{table}[]
    \centering
    \caption{Ablation study of our components, replaced with the general approach of \cite{Ho2018Submaps}, as mean and standard deviation of 10 runs in the maze at Moderate drift.}
    \begin{tabular}{lcc}
        \toprule
        \parbox{0.45\columnwidth}{\centering Method\\(Local Area, Submap Frontiers)} &
            \parbox{0.2\columnwidth}{\centering Exploration 50\% [min]} &
            \parbox{0.2\columnwidth}{\centering Exploration 90\% [min]}  \\
        \midrule
        GLocal (Ours, Ours) & $\mathbf{2.32 \pm0.53}$ & $\mathbf{7.85 \pm1.74}$ \\
        No Local Area (\cite{Ho2018Submaps},  Ours) & $3.23 \pm0.90$ & $8.39 \pm1.59$ \\
        No Submap Frontiers (Ours, \cite{Ho2018Submaps}) & $2.48 \pm0.78$ & $8.98 \pm1.37$ \\
        Global Submaps (\cite{Ho2018Submaps},  \cite{Ho2018Submaps}) & $3.66 \pm0.80$ & $9.27 \pm1.03$ \\
        \bottomrule
    \end{tabular}
    \label{tab:ablation}
    \vspace{-5mm}
\end{table}

\subsection{Computational Complexity}
\begin{figure}[]
\centering
\includegraphics[width=\linewidth]{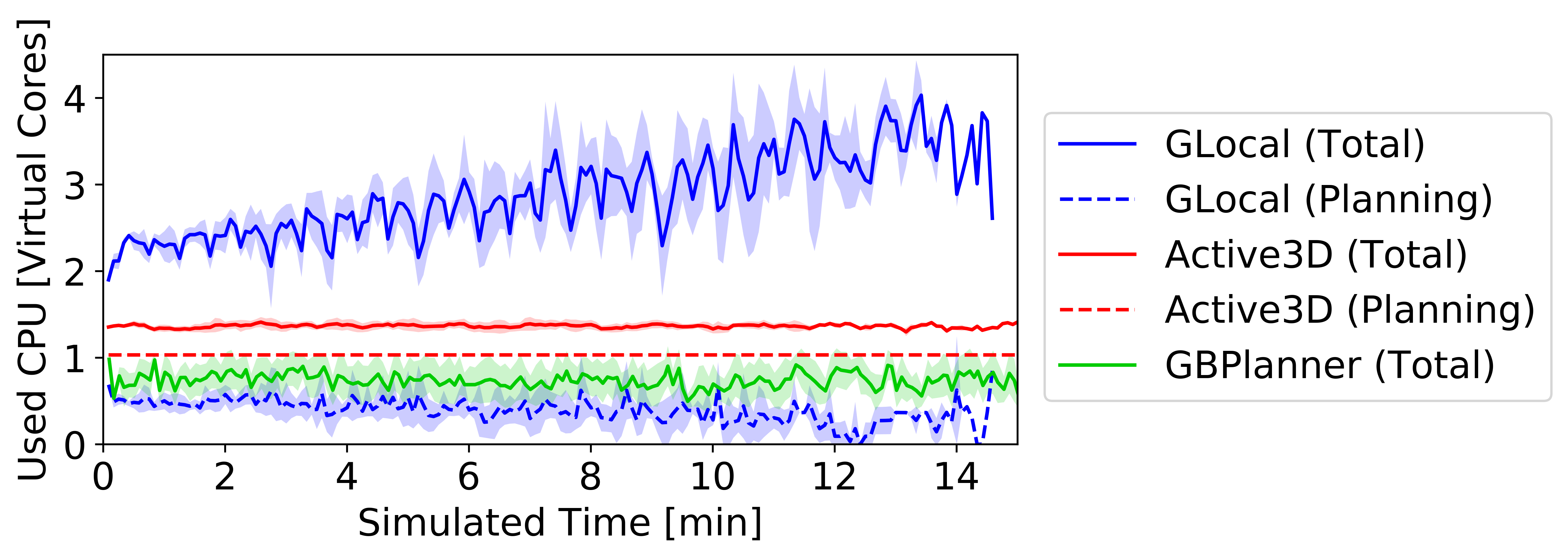} 
\caption{CPU consumption over time as mean and standard deviation of 10 runs in the Maze at Moderate drift. Similar to Active3D and GBPlanner, the majority of our system runs in constant time on two virtual cores. An additional, linearly scaling map optimization part can also be distributed over longer periods of time, making the system suitable for operation on-board a mobile platform.}
\label{fig:cpu_planner}
\vspace{-2mm}
\end{figure}

A bearable computational load is highly relevant for operation on-board a mobile platform
Thus, the average CPU consumption of each system is shown in Fig.~\ref{fig:cpu_planner}. Data is collected on the simulation computer using an Intel i9 9900K.
While GBPlanner is most efficient, Active3D shows constant consumption at an increased level.
Similarly, GLocal runs on two threads in constant time with additional map pose graph optimization.
The latter scales linearly with the number of submaps and is responsible for periodic bursts when the pose graph is being optimized.
The map optimization could be distributed over longer periods of time to accommodate systems with less powerful hardware \cite{reijgwart2020voxgraph}.

\begin{figure}[]
\centering
\includegraphics[width=0.48\linewidth]{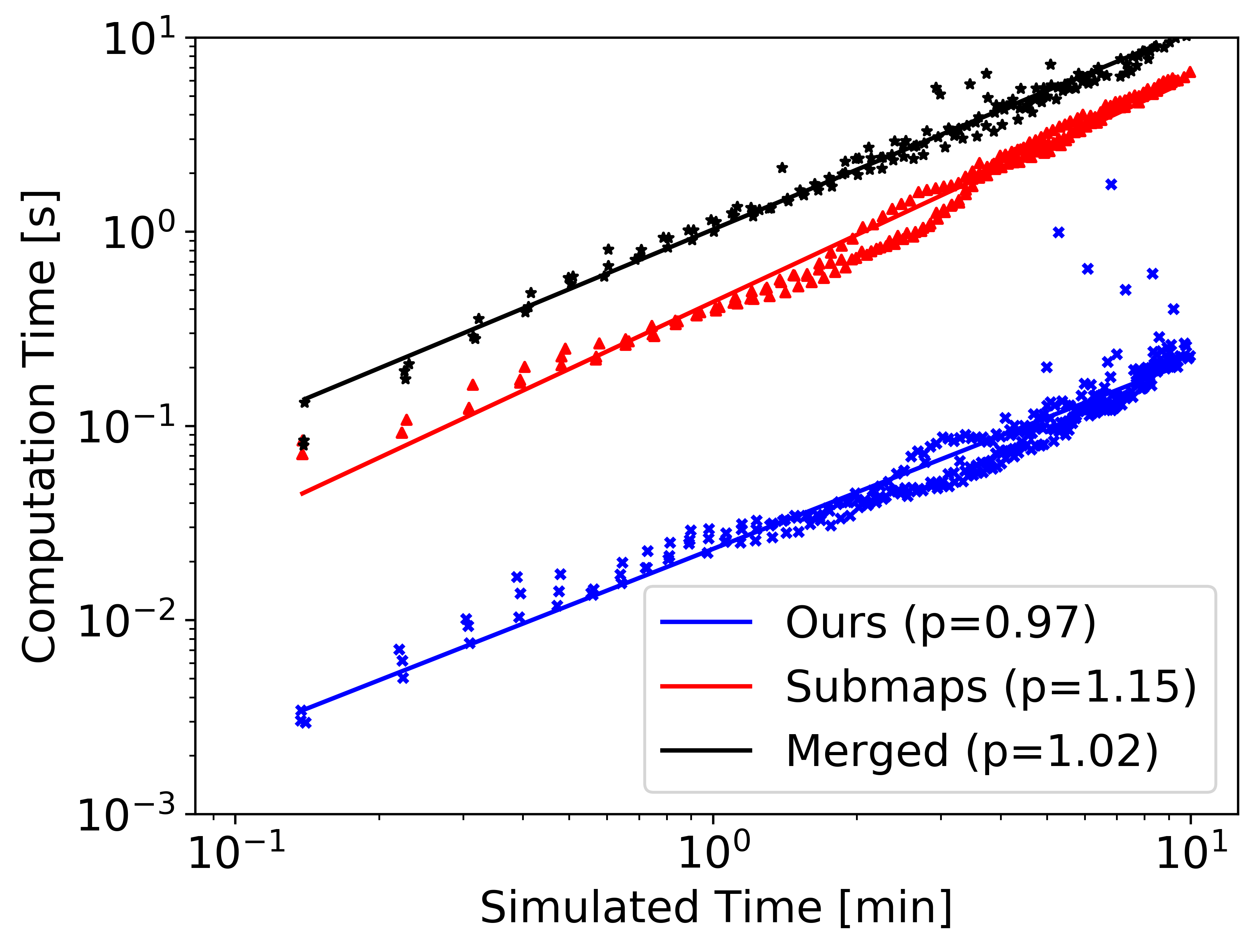} 
\hfill
\includegraphics[width=0.48\linewidth]{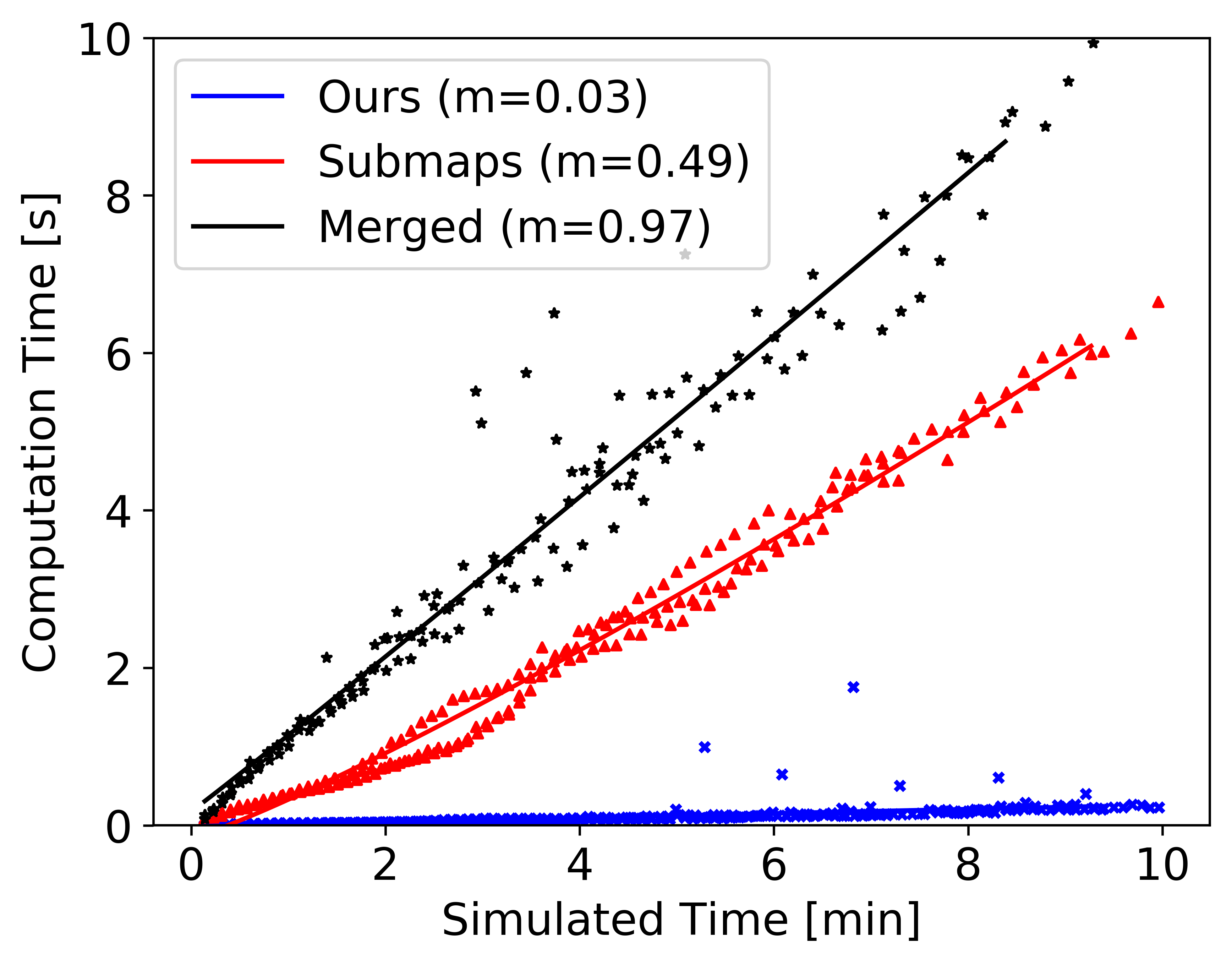}
\caption{Computation times to detect global frontiers against exploration progress. Power analysis (left) shows that the merged and our approach scale linearly with time, whereas Submap scales super-linearly. The fitted computation times (right) indicate a $\times30$ speedup of our submap frontier approach over merged. }
\label{fig:cpu_frontiers}
\vspace{-5mm}
\end{figure}

The computation cost to find global frontiers as exploration progresses is detailed in Fig.~\ref{fig:cpu_frontiers}. 
We compare the presented submap frontier approach (Ours) against a global frontier search through the current submap collection, as proposed in \cite{Ho2018Submaps} (Submap). 
A third option is to merge all submaps into a global map that can then be queried in $O(1)$ for global frontier search (Merged). 
Power analysis (Fig. \ref{fig:cpu_frontiers}, left) suggests that the methods scale $\propto mt^p$, with $t$ being exploration time. 
Fitting for $p$ shows that Ours and Merged scale $O(t)$, whereas Submap scales super-linear.
Similarly, fitting for the multiplicative factors $m$ (Fig.~\ref{fig:cpu_frontiers}, right) indicates a speedup of $\times30$ of our method over Merged.

\subsection{Field Experiments}
\begin{figure}[]
\centering
\includegraphics[width=0.38\linewidth]{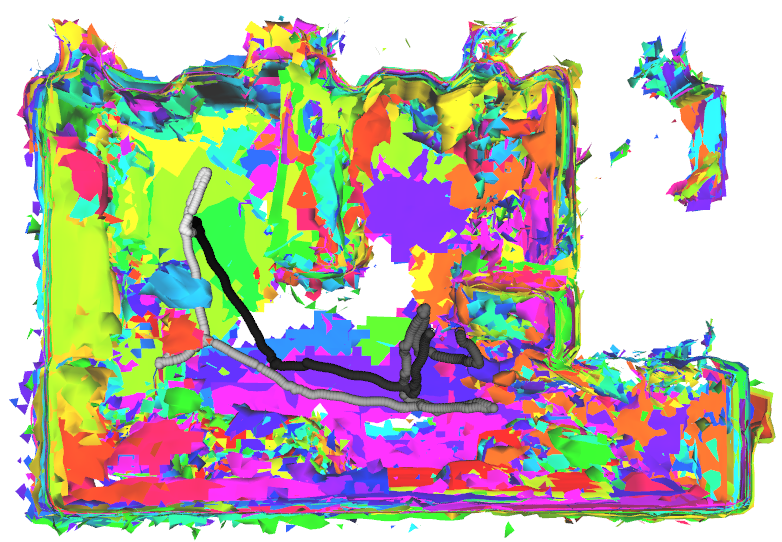} 
\hfill
\includegraphics[width=0.6\linewidth]{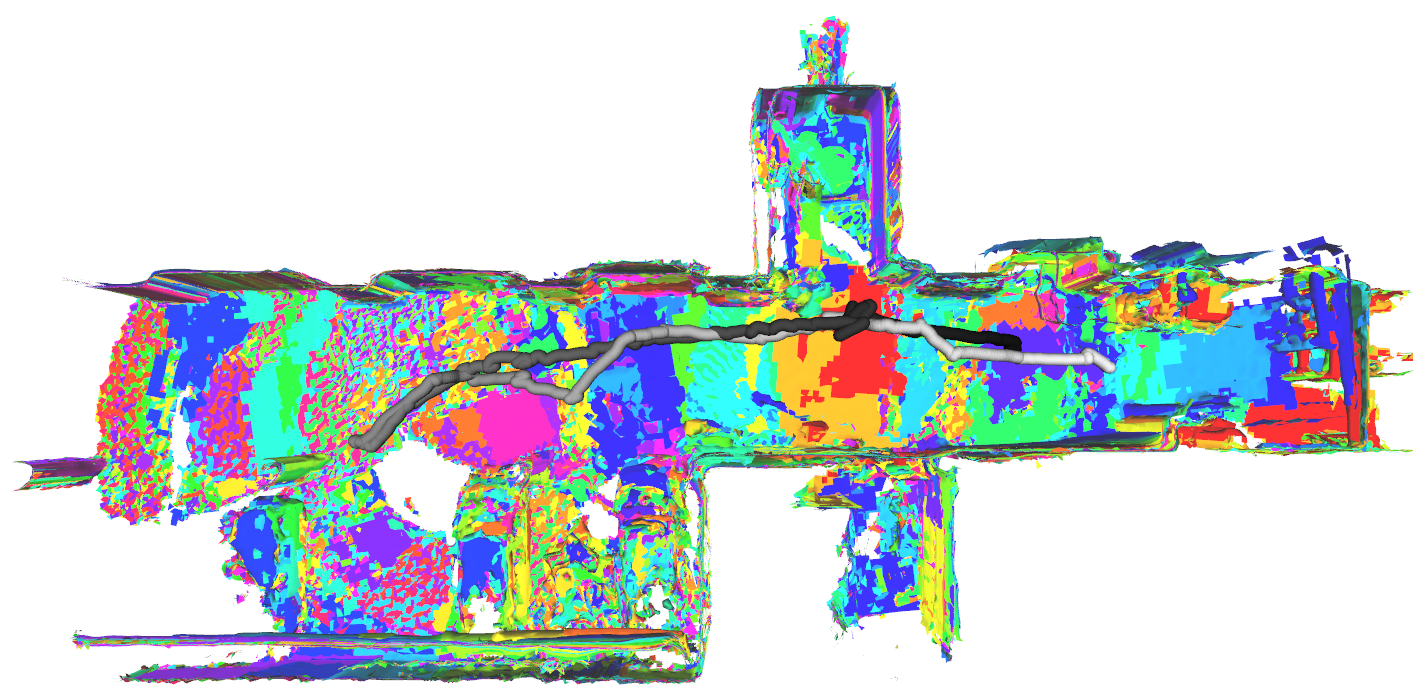}
\\
\includegraphics[width=0.38\linewidth]{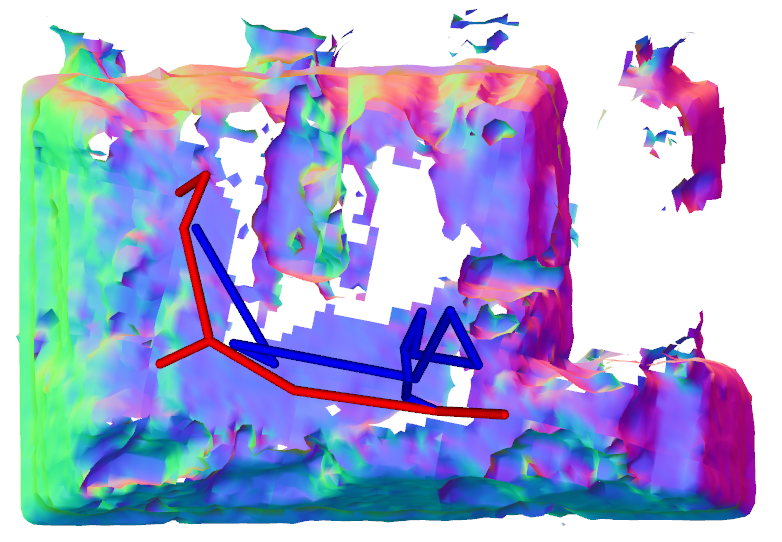} 
\hfill
\includegraphics[width=0.6\linewidth]{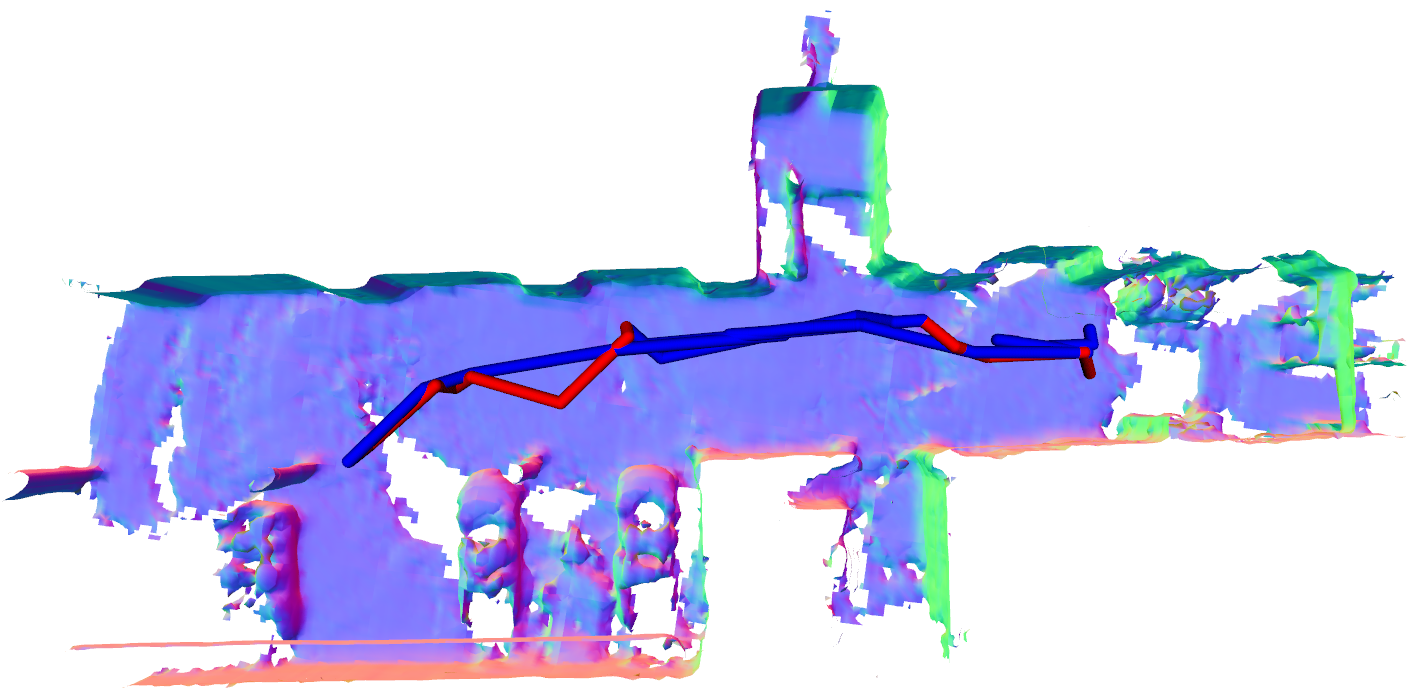}
\caption{Experiments on a fully autonomous MAV highlight GLocal's performance. Top: Submaps and executed path from start (white) to finish (black). Bottom: Combined reconstruction and planned local (red) and global (blue) path, obtained in an indoor scene (left) and a large scale underground parking area (right).}
\label{fig:real_exp}
\vspace{-6mm}
\end{figure}

To further validate our method, we perform exploration experiments in two challenging environments with a MAV using only on-board sensing and computing. 
We use \cite{rovio} for visual-inertial state estimation.
The results are shown in Fig.~\ref{fig:real_exp}. 
The left and right columns correspond to an indoor room of $12\times8\times4\SI{}{m}$ split into 3 compartments, and an underground parking area of $42\times16\times4.5\SI{}{m}$, respectively.
The top row shows the individually colored submaps and the executed path, colored from start (white) to finish (black).
The bottom row shows the combined reconstruction colored by surface normals and the planned local (red) and global (blue) paths.

In both cases, GLocal shows desirable behavior, starting by planning locally and peeking into corners and behind cars and pillars. The global planner ensures that missed areas are revisited and leads to complete exploration. Only a small number of holes in the floor remain due to sensing restrictions on the used platform. Due to alignment of the submaps to compensate for drift, a high quality map with straight and perpendicular walls is obtained. 
All processes ran solely on-board an Intel NUC (i7 8650U), where GLocal used an average of 2.62 virtual CPU-cores, 0.67 of which for planning.


\section{Conclusions}
In this work, an approach that combines multiple layers of both mapping and planning to enable safe and efficient volumetric exploration of large scale environments when subject to odometry drift was proposed.
We show that a submap-based, dense map representation can be leveraged to provide the temporally and spatially local information required for efficient local exploration under drift in constant lookup time, while still being able to accommodate past pose corrections in the global map.
Furthermore, we present a two stage planning approach that leverages the submap structure to efficiently compute global frontiers in a changing 3D volumetric map.
The proposed system was thoroughly evaluated in extensive, high-fidelity simulation, and shown to outperform state of the art methods in terms of performance while enabling safety and robustness when subjected to odometry drift.
Experiments on a real MAV further validate the method.
Our \emph{GLocal} system and the simulator are made available as open source.

\section*{ACKNOWLEDGMENT}
We are grateful for Christian Lanegger and Alexander Millane's support during the robot experiments.


{\small
\bibliographystyle{IEEEtran}
\bibliography{IEEEfull,references}
}


\end{document}